\documentclass[11pt,twocolumn]{article}

\usepackage{acl}

\usepackage{times}
\usepackage{latexsym}
\usepackage{booktabs}
\usepackage{ulem}
\usepackage{amsmath}
\usepackage{subcaption}
\usepackage{longtable}
\usepackage{multirow}
\usepackage{placeins}
\usepackage{tikz}
\usetikzlibrary{positioning,fit,arrows.meta}
\usepackage{needspace}
\usepackage{amsfonts}
\usepackage{microtype}
\usepackage{inconsolata}
\usepackage{graphicx}
\usepackage{algorithm}
\usepackage{algorithmic}
\usepackage{tabularx}
\newcommand{\prompt}{\mathcal{P}}

\usepackage[T1]{fontenc}
\usepackage[utf8]{inputenc}

\title{Zero-Shot Visualization: Exploring Text Corpora with User-Prompted Axes}
\date{}

\author{Arnau Bueno Tricas, Jose A. Rodriguez-Serrano\\
  Universitat Ramon Llull Esade \\ Barcelona, Spain \\
}

\begin{document}
\maketitle

\begin{abstract}
We study the application of large language models (LLMs) to the visual exploration of textual corpora. We introduce \textit{zero-shot visualization} (ZSV), a task in which users specify concepts in natural language and documents are mapped onto the corresponding concept axes for visualization.
Building a ZSV system of practical value is non-trivial, as
it requires choices at the intersection of feature functions, efficient implementation tradeoffs, and pre/post-processing decisions affecting visualization quality. 
To that end, we establish a benchmark that compares methods spanning embedding similarity, direct semantic judgments, and conditional likelihood estimation in this setting. Across multiple datasets and use cases we evaluate the properties of different scoring methods and design choices in terms of semantic faithfulness, score fidelity, and computational cost. Our results identify that scoring based on next-token probabilities offers the strongest practical trade-off among the evaluated methods. We further apply this approach to unlabeled corpora to examine its behavior in realistic exploratory settings. These experiments highlight additional design considerations, including the use of graded axes together with binary relevance filtering, and reveal a compositional sentiment bias in off-topic documents. Based on these findings, we provide practical guidelines for constructing end-to-end ZSV baselines.
\end{abstract}

\section{Introduction}

\begin{figure}[!t]
  \includegraphics[width=\linewidth]{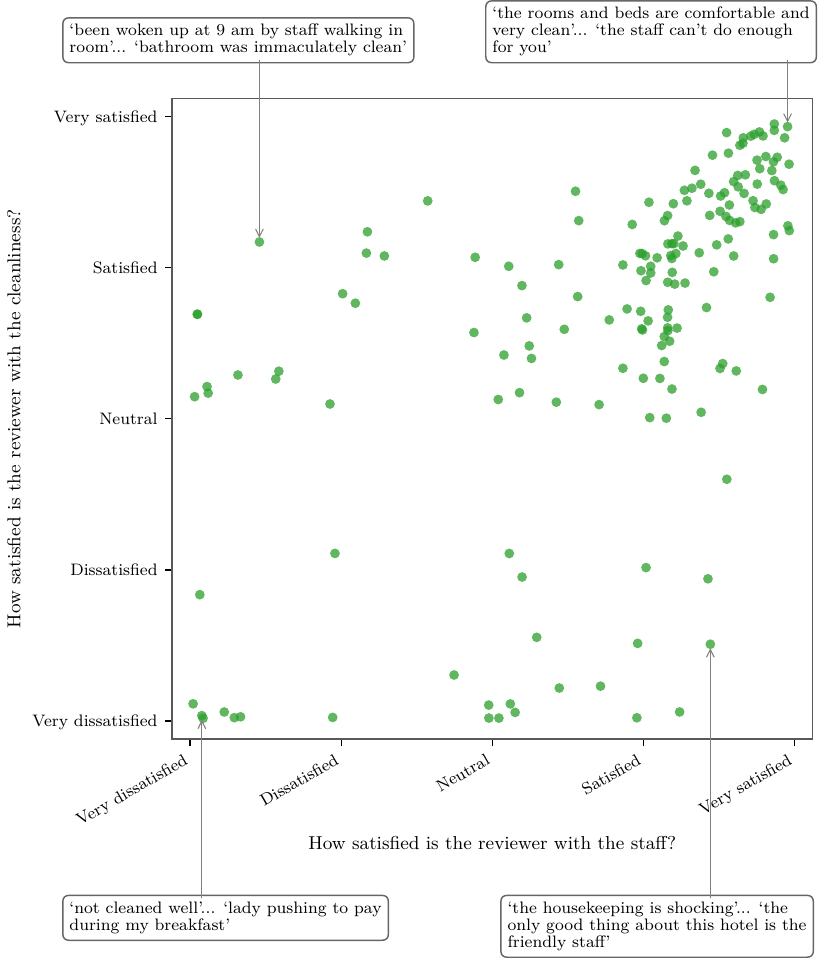}
  \caption{ZSV on an unlabeled corpus of hotel reviews~\citep{liu2017515k}. Reviews are positioned along two user-defined graded axes (``How satisfied is the reviewer with the staff?'' and ``How satisfied is the reviewer with cleanliness?''; Section~\ref{sec:graded}), filtered to reviews discussing both. Corner examples ground the geometry in text.}
  \label{fig:zsv_teaser}
\end{figure}

Large language models (LLMs) are increasingly used as general-purpose interfaces
for interacting with data~\cite{chen2024}.
In data exploration, much prior work has adopted conversational interaction patterns, in which users pose natural-language
queries and receive textual answers or summaries~\citep{weng2025,ouyang2025}. While effective for guided
analysis and question answering, this paradigm treats the LLM primarily as an
external reasoning engine, rather than as a component that directly shapes the
representations used for exploration.

In parallel, classical data exploration and visualization methods rely on
explicit feature representations to structure datasets, identify patterns, and
support interactive analysis~\citep{maaten2008,mcinnes2018umap}. Bridging these paradigms raises a natural question:
can the semantic knowledge encoded in LLMs be used directly to construct
feature spaces for exploration, instead of only mediating user--system
interaction?

In this work, we focus on textual datasets and propose Zero-Shot Visualization (ZSV): the task of exploring text corpora through natural-language-defined axes without task-specific training. A user defines axes by writing natural-language concepts (e.g., ``Does the review talk about the hotel location''), and a scoring function measures how closely each text aligns with each concept. These scores act as coordinates in a low-dimensional space, enabling projection, visualization, and clustering along axes that users specify and refine interactively. 
Any practical realization of ZSV faces a number of requirements. First, 
any function that maps a document to a concept must be \textit{faithful}, that is, it needs to make sense semantically. Second, it needs to be computationally efficient: we envision a user specifying the axes and getting the result after a few seconds. Third, it should run on commodity hardware, for natural integration with other common tools used for data exploration (e.g. should be locally invoked from within a Jupyter notebook). 

Therefore, the design choices for a ZSV system are not trivial. Traditional semantic projection methods like t-SNE~\cite{maaten2008} or UMAP~\cite{mcinnes2018umap} achieve a similar result, but do not map concepts to axes explicitly. Verbalizers that map texts to concepts by prompting LLMs~\cite{xiong2024can} are direct candidates but their use on commodity hardware is not efficient. Embedding methods~\cite{reimers2019sentencebertsentenceembeddingsusing, lassance2024spladev3newbaselinessplade} are faster but, as will be shown, the scores are inferior and not properly scaled to be used as axes.

Our contributions are threefold: (a) the {\bf framework} (Section~\ref{sec:method}; formalizing ZSV and identifying scoring families that satisfy its requirements), (b) a {\bf visualization-first benchmark and study of exploratory cases} (Section~\ref{sec:experiments}; systematically comparing scoring methods for interactive corpus exploration across multiple datasets and use cases), and (c) {\bf practitioner guidance} (Section~\ref{sec:practitioner}; practical guidelines in terms of method choice, LLM choice, scaling, inference cost, and pre-processing).

\section{Related Work}
\label{sec:related_work}

Recent work has investigated natural language as an interface for data exploration and visualization, allowing analytical intents or visualization requests to be expressed through free-form text. This includes producing charts given written instructions~\cite{belouadi2025}, grounding instructions with tabular data~\cite{shi2025,wang2024}, and using agentic frameworks to deal with multiple data sources~\cite{ouyang2025} or to extend to general exploratory data analysis tasks~\cite{weng2025}, and benchmarks such as VisEval~\cite{chen2024}. These systems focus on {\it tabular data} and translate language into visualization specifications over existing data attributes, rather than defining new semantic dimensions.

Another body of work has focused on chart understanding, summarization or visual question answering~\cite{kafle2018,masry2022,liu2023,hoque2025}. This line of work assumes a fixed visualization and focuses on interpretation, whereas our work addresses how semantic axes are constructed in the first place.

The zero-shot extraction of labels, classes or concepts, or their scores, is not only an intrinsic property of LLMs (when prompted adequately)~\cite{bucher2024,gretz2023}, but also the subject of an ongoing body of related work on verbalizers~\cite{zhao2023,xiong2024can}, or zero-shot classification~\cite{wang2023,kumar2023,zhang2024}, including the use of small models~\cite{lepagnol2024}. In extensions to multimodal data, zero-shot learning has shown tremendous success in zero-shot, open vocabulary image classification~\cite{fu2019,radford2021}.

However, these works target label generation rather than visualization, and any byproduct projections are unsupervised, not determined by written specifications~\cite{wang2023}. Closer to ZSV, recent work uses LLMs to extract or score natural-language concepts over text corpora, either by inducing concepts from the data~\cite{lam2024concept,bravansky2025datasetfeaturizationuncoveringnatural} or by embedding concept scoring within a trained classifier~\cite{sun2024craftinglargelanguagemodels}. ZSV instead takes concepts as exogenous user input and systematically benchmarks scoring families for interactive corpus exploration, rather than committing to a single fixed scorer.

Finally, classical projection methods such as t-SNE~\cite{maaten2008}, UMAP~\cite{mcinnes2018umap} or visualization methods for topic models~\cite{chaney2012} are commonly employed to visualize textual datasets. While practical and well-established, these are unsupervised techniques unable to assign concept names or scores. A closer precursor is \cite{6634124}, whose explainers craft user-aligned, namable projection axes for high-dimensional data exploration. Unlike ZSV, the approach requires example-based annotations and depends on a pre-existing numeric feature representation rather than scoring raw text. Other work explored interactive improvements to unsupervised methods~\cite{endert2012,iwata2013}, allowing limited steering but lacking natural-language, on-demand semantic axis definition.

\section{Zero-Shot Visualization Framework}
\label{sec:method}

\subsection{Problem Setup}
\label{sec:problem_setup}

Let $\mathcal{X} = \{x_1,\dots,x_N\}$ be a corpus of texts and
$\mathcal{C} = \{c_1,\dots,c_K\}$ a set of user-defined semantic concepts,
each expressed as a natural-language requirement (e.g., ``This question is
about astronomy''). No labeled data linking texts to concepts is assumed.

The goal of Zero-Shot Visualization is to map each text $x$ to a
$K$-dimensional semantic representation
\[
\mathbf{s}(x) = \big(s(x,c_1), \dots, s(x,c_K)\big),
\]
where $s(x,c_k) \in [0,1]$ denotes the estimated alignment between text $x$
and concept $c_k$ computed in a zero-shot manner.

Each concept defines a semantic axis in an explicit, human-interpretable
feature space. Unlike latent embedding spaces, this representation is fully
specified by natural-language concepts chosen by the user.

Once computed, concept activations can be used directly for visualization
(e.g., 2D or 3D projections), clustering, and downstream modeling or decision-making tasks, without retraining or task-specific fine-tuning.

\subsection{Design Requirements}
\label{sec:requirements}

As a visualization-first benchmark, ZSV is designed for interactive corpus exploration: an analyst defines axes on the fly and inspects the resulting concept space. This use case imposes three concrete requirements on any candidate scoring method:

\begin{enumerate}
    \item \textbf{Bounded, faithful scores.} Scores must lie in a fixed range ($[0,1]$ in our case) so that distances and thresholds are meaningful across texts and concepts. When ground-truth labels are available, scores should accurately recover the corresponding semantic structure.
    \item \textbf{Near real-time inference.} The analyst defines and refines axes during exploration, so scoring must be fast enough to keep the analysis loop responsive on user-prompted concepts.
    \item \textbf{Commodity hardware.} The pipeline must run on a standard consumer-grade machine without specialized infrastructure or remote inference services.
\end{enumerate}

For these reasons, all experiments in this paper use small open-source models that fit on a standard personal computer. Open-source models additionally ensure reproducibility, allow the pipeline to be shared without API dependencies, and expose the next-token probabilities required by several scoring methods, which are not made available by all proprietary APIs.

\subsection{Zero-Shot Scoring Functions}
\label{sec:scoring}

ZSV is compatible with any method producing a bounded alignment score $s(x,c)$ between a text and a natural-language concept, including embedding similarity, zero-shot classification, and LLM-based scoring.

We analyse baseline and state-of-the-art methods that
can be used as scorers and 
organize them into two main families. Implementation details for every scorer are listed in Appendix~\ref{app:implementation-prompts}.

Retriever-based scorers compute alignment via similarity in a fixed lexical
or semantic vector space, while LLM-based scorers directly estimate
semantic alignment through language model inference.
Classical lexical relevance functions such as BM25~\citep{robertson2009probabilistic}
are not considered, as their scores are unbounded and scale-dependent, making
them unsuitable for direct geometric comparison.

% -------------------------------------------------
\subsubsection{Retriever-Based Scorers}
\label{sec:retrievers}

Representation-based retrievers estimate alignment by embedding both texts
and concepts into a shared vector space and computing similarity via a fixed
metric.
These methods are deterministic, computationally efficient, and scalable,
making them well-suited as coarse semantic approximations and as first-stage
scorers in hybrid pipelines.

\paragraph{Sparse Lexical Retriever (SLR).} 
Alignment is computed by representing both the text and the concept description
as high-dimensional sparse vectors over the vocabulary and measuring their cosine
similarity.
In our implementation, this baseline retriever is instantiated using a standard TF--IDF
representation.
The resulting scores are fully deterministic and interpretable, but rely on
exact lexical overlap and therefore fail to capture paraphrastic or implicit
semantic relations.

\paragraph{Dense Neural Retriever (DNR).}
Texts and concepts are encoded into a shared low-dimensional semantic space using
a pretrained neural sentence encoder.
Let $\phi(x), \phi(c) \in \mathbb{R}^d$ denote L2-normalized dense embeddings.
Alignment is computed as cosine similarity,
\[
s(x,c) = \frac{1 + \langle \phi(x), \phi(c) \rangle}{2},
\]
yielding bounded scores in $[0,1]$.
In contrast to sparse lexical retrieval, this formulation captures semantic
similarity beyond exact word overlap while remaining computationally efficient.
In our experiments, we instantiate this retriever using a MiniLM-based
Sentence-BERT model \citep{wang2020minilmdeepselfattentiondistillation,reimers2019sentencebertsentenceembeddingsusing},
chosen for its favorable trade-off between embedding quality and computational
efficiency \citep{muennighoff2023mtebmassivetextembedding}.

\paragraph{Sparse Neural Retriever (SNR).}
Sparse neural retrievers combine lexical matching with learned term reweighting
and expansion.
Texts and concepts are represented as high-dimensional sparse vectors over the
vocabulary, with nonzero entries predicted by a neural model.
Alignment is computed using cosine similarity in this sparse vocabulary space,
yielding bounded scores comparable to other retriever classes.
Compared to sparse lexical retrieval, this formulation enables learned lexical
generalization while retaining interpretability and retrieval efficiency.
In our experiments, we instantiate this class using a SPLADE model
\citep{lassance2024spladev3newbaselinessplade}.

\subsubsection{LLM-Based Scorers}
\label{sec:llm_scorers}

Judgment-based scorers use a large language model to directly assess whether a
text satisfies a given concept.
Unlike retrievers, these methods do not rely on a fixed representation space,
but instead produce alignment estimates through conditional generation or
probabilistic inference.
While semantically expressive, they are substantially more expensive to apply
exhaustively.

\paragraph{Direct Rating Scoring (DR).}
In the direct rating approach, a large language model (LLM) is prompted to assess
whether text $x$ satisfies concept $c$ and return a single scalar score on a fixed
numeric scale.
The returned score is linearly normalized to $[0,1]$ to yield $s(x,c)$.
While semantically expressive, this method may be sensitive to prompt formulation
and discretization effects induced by numeric generation.

\paragraph{Probabilistic Zero-Shot Scoring (PZS).}
To obtain well-scaled and comparable scores, we formulate zero-shot inference as a
binary probabilistic decision derived from the LLM's next-token distribution. Let $\pi(\cdot \mid \prompt(x,c))$ denote the
language model's next-token distribution under a binary prompt
$\prompt(x,c)$ that queries whether text $x$ satisfies concept $c$.
Following our implementation, we extract the probability mass assigned to an
affirmative response by summing the probabilities of a small set of token
variants corresponding to \texttt{True}:
\begin{equation}
\label{eq:p_true}
p_{\mathrm{T}}(x,c)
= \sum_{t \in \mathcal{T}_{\mathrm{true}}}
\pi\!\left(t \mid \prompt(x,c)\right).
\end{equation}

Analogously, we compute $p_{\mathrm{F}}(x,c)$ for negative responses and define
the normalized alignment score $s(x,c) = \Pr(\mathrm{True} \mid x,c)$, given by
\begin{equation}
\label{eq:alignment_score}
s(x,c)
= \frac{p_{\mathrm{T}}(x,c)}
       {p_{\mathrm{T}}(x,c)+p_{\mathrm{F}}(x,c)+\varepsilon}
\in [0,1],
\end{equation}
where $\varepsilon>0$ is a small constant used for numerical stability

\paragraph{Explanation-Marginalized Scoring (EM).}
To reduce sensitivity to individual reasoning traces, we estimate alignment
by marginalizing predictions over multiple sampled explanations, following
\citep{becker2024cycles}. Given a text $x$, we sample $N$ short reasoning
traces $e_i \sim p(e \mid x)$, $i=1,\dots,N$. For each explanation $e_i$, we
compute an entailment probability $q_i$ (whether the explanation is
supported by the text) and an alignment probability $a_i$ (whether the
concept holds given both the text and the explanation):
\begin{align}
q_i &= \Pr(\mathrm{True} \mid \prompt_{\mathrm{ent}}(x,e_i)), \\
a_i &= \Pr(\mathrm{True} \mid \prompt_{\mathrm{req}}(x,c,e_i)).
\end{align}
Both probabilities use the same normalized True/False next-token mass as
PZS (Eqs.~\eqref{eq:p_true}--\eqref{eq:alignment_score}). Forming
normalized entailment weights $w_i = q_i / (\sum_{j} q_j + \varepsilon)$,
we return the marginal alignment score
\begin{equation}
s_{\mathrm{marg}}(x,c) = \sum_{i=1}^{N} w_i\, a_i \;\approx\;
\mathbb{E}_{e \sim p(e\mid x)}\!\left[a(x,c,e)\right],
\end{equation}
at $3N$ prompts per pair, yielding a more stable estimate of semantic
alignment at substantially higher cost than PZS.

\subsubsection{Hybrid Coarse-to-Fine Scoring}
\label{sec:hybrid}
Some LLM-based scorers are prohibitively expensive to apply exhaustively over large corpora. We adopt a hybrid coarse-to-fine strategy: a fast retriever first scores all text--concept pairs, then a higher-fidelity LLM scorer re-evaluates a restricted subset (top-$k$, percentile, or threshold); unselected pairs retain their coarse scores. Final quality is limited by the first stage. We implement this with EM as the reranker and call the resulting configuration \textbf{EM-Hybrid}.

\section{Experiments}
\label{sec:experiments}

We evaluate ZSV across two complementary regimes that reflect common text analysis workflows: (i) tagged corpora, where labels enable direct assessment of semantic faithfulness; and (ii) untagged corpora, where the goal is exploratory discovery without supervision.

\subsection{Tagged Corpora: Semantic Faithfulness}
\label{sec:tagged}

We evaluate ZSV on tagged text corpora, where
ground-truth semantic categories are available. The objective is not to optimize
classification accuracy, but to assess semantic faithfulness: whether
zero-shot alignment with natural-language concepts recovers known label
structure without training, calibration, or representation tuning.

\paragraph{Datasets and concepts.}
We use two benchmark datasets with human-annotated categories: MMLU, a collection
of multiple-choice questions spanning academic subjects \citep{hendryckstest2021},
and Banking77, a dataset of customer service queries labeled by intent
\citep{Casanueva2020}. For each dataset, we define one natural-language requirement
per ground-truth label as a short declarative statement describing the category.

\paragraph{Quantitative results.}
Each text is mapped to a concept vector by scoring its alignment with
all requirements using the scorers in Section~\ref{sec:scoring}.
Semantic faithfulness is evaluated quantitatively, treating each
score as a one-vs-all classifier and computing AUROC, Average Precision
(AP), and Stratified Brier (per-class Brier averaged over positives and
negatives, so a constant predictor scores $0.5$). High AUROC and AUC indicate alignment with the semantic intent of the task, low Brier score additionally indicates good scaling and ground-truth alignment -- in the sense that the score is close to 1 or 0, as expected, for these binary concepts with known labels.
%; and (ii) qualitatively,
%via low-dimensional projections of the concept space. 
%Tables report means over MMLU subjects ($n=57$) and Banking77 intents ($n=77$).

%\paragraph{Quantitative results.}
Table~\ref{tab:tagged_results} reports results on MMLU and Banking77. Tables report means over MMLU subjects ($n=57$) and Banking77 intents ($n=77$).
Unless stated otherwise, all LLM-based scorers use Mistral-7B-Instruct-v0.3 as the reference LLM. %; cross-model consistency at the 3B--9B scale is verified in Table~\ref{tab:robustness}. 
For EM-Hybrid, SLR serves as the first-stage retriever and EM reranks the top $10\%$ ($N=3$); standalone EM is omitted due to its compute cost. PZS leads on all metrics; DNR follows but with notably worse ground-truth alignment (Brier).

%%%%

\begin{table}[!t]
\centering
\footnotesize
\setlength{\tabcolsep}{3pt}
\renewcommand{\arraystretch}{1.1}
\begin{tabular}{lcccccc}
\toprule
 & \multicolumn{3}{c}{\textsc{MMLU}} & \multicolumn{3}{c}{\textsc{Banking77}} \\
\cmidrule(lr){2-4} \cmidrule(lr){5-7}
\textbf{Scorer}
 & \textbf{AUROC} & \textbf{AP} & \textbf{Brier}$\downarrow$
 & \textbf{AUROC} & \textbf{AP} & \textbf{Brier}$\downarrow$ \\
\midrule
\multicolumn{7}{l}{\textit{Retriever-Based Scorers}} \\
SLR & $0.55$ & $0.07$ & $0.48$ & $0.84$ & $0.30$ & $0.35$ \\
DNR & $0.85$ & $\underline{0.20}$ & $\underline{0.22}$ & $\mathbf{0.95}$ & $\mathbf{0.55}$ & $0.21$ \\
SNR & $0.79$ & $0.17$ & $0.45$ & $\underline{0.94}$ & $\underline{0.54}$ & $\underline{0.20}$ \\
\midrule
\multicolumn{7}{l}{\textit{LLM-Based Scorers}} \\
DR  & $\underline{0.86}$ & $0.15$ & $0.29$ & $0.69$ & $0.06$ & $0.36$ \\
PZS & $\mathbf{0.93}$ & $\mathbf{0.42}$ & $\mathbf{0.14}$ & $\mathbf{0.95}$ & $\mathbf{0.55}$ & $\mathbf{0.11}$ \\
\midrule
\textit{EM-Hybrid} & $0.57$ & $0.11$ & $0.44$ & $0.84$ & $0.42$ & $0.27$ \\
\bottomrule
\end{tabular}
\caption{Semantic faithfulness on tagged MMLU and Banking77 (validation splits). \textbf{Bold} marks the best result per metric; \underline{underline} marks the second-best.}
\label{tab:tagged_results}
\end{table}

\paragraph{Computational Cost.}
To quantify the computational cost of different alignment strategies, we ran a
controlled timing experiment on the abstract algebra subset of MMLU (116
texts), evaluated against a single requirement. Under this setting,
retriever-based scorers complete in milliseconds (SLR: 68\,ms; DNR: 296\,ms; SNR:
341\,ms), while LLM-based methods incur higher costs (PZS: 2.9\,s; DR: 3.0\,s; EM:
54.0\,s). The substantially higher runtime of EM reflects its linear scaling in
the number of sampled explanations ($3N$ prompts per pair), whereas PZS and DR
rely on a single prompt. Taken together, these results motivate PZS as the strongest practical compromise
among all methods considered.

\paragraph{Robustness across LLMs.}
To verify that PZS results are not an artifact of using Mistral-7B-Instruct-v0.3,
we replicate the 1-vs-all evaluation across four comparable open-weight models
(Qwen2.5-3B-Instruct, Mistral-7B-Instruct-v0.3, Llama-3.1-8B-Instruct, Gemma-2-9B-it;
Table~\ref{tab:robustness}). Results are consistent across models, indicating
that the reported behavior is not specific to a single LLM.

\begin{table}[!t]
\centering
\footnotesize
\setlength{\tabcolsep}{2pt}
\renewcommand{\arraystretch}{1.1}
\begin{tabular}{lcccccc}
\toprule
 & \multicolumn{3}{c}{\textsc{MMLU}} & \multicolumn{3}{c}{\textsc{Banking77}} \\
\cmidrule(lr){2-4} \cmidrule(lr){5-7}
\textbf{Model}
 & \textbf{AUROC} & \textbf{AP} & \textbf{Brier}$\downarrow$
 & \textbf{AUROC} & \textbf{AP} & \textbf{Brier}$\downarrow$ \\
\midrule
Qwen-3B    & $0.93$ & $\underline{0.42}$ & $0.40$ & $0.94$ & $0.52$ & $0.28$ \\
Mistral-7B & $0.93$ & $\underline{0.42}$ & $\underline{0.14}$ & $\underline{0.95}$ & $\underline{0.55}$ & $\underline{0.11}$ \\
Llama-8B   & $\underline{0.94}$ & $\underline{0.42}$ & $0.22$ & $0.92$ & $0.45$ & $0.22$ \\
Gemma-9B   & $\mathbf{0.96}$ & $\mathbf{0.50}$ & $\mathbf{0.10}$ & $\mathbf{0.96}$ & $\mathbf{0.63}$ & $\mathbf{0.09}$ \\
\bottomrule
\end{tabular}
\caption{Cross-LLM PZS consistency. \textbf{Bold} marks the best result per metric; \underline{underline} marks the second-best.}
\label{tab:robustness}
\end{table}

\paragraph{Qualitative analysis.}

\begin{figure}[!t]
  \centering
  \begin{subfigure}{0.48\linewidth}
    \includegraphics[width=\linewidth]{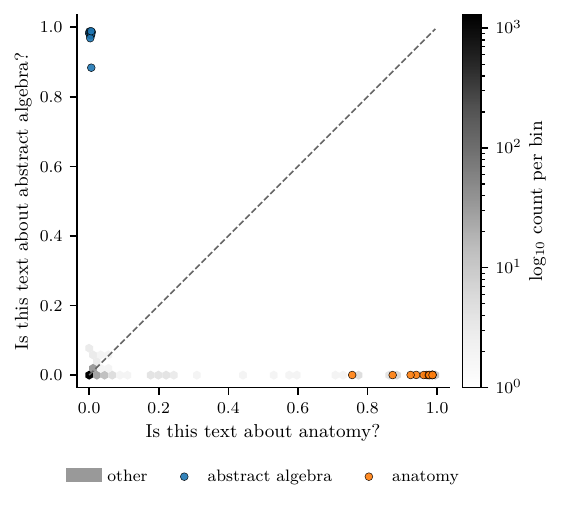}
    \caption{Orthogonal}
    \label{fig:mmlu_orthogonal_hexbin}
  \end{subfigure}
  \hfill
  \begin{subfigure}{0.48\linewidth}
    \includegraphics[width=\linewidth]{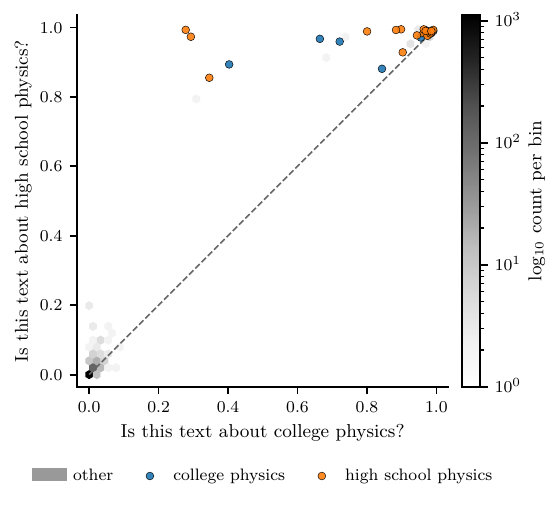}
    \caption{Related}
    \label{fig:mmlu_related_hexbin}
  \end{subfigure}
  \caption{MMLU concept space using the PZS scorer. Grey hexagons show log-scaled density of texts outside the highlighted categories; colored points denote highlighted subjects.}
  \label{fig:mmlu_hexbin_panel}
\end{figure}

Figure~\ref{fig:mmlu_hexbin_panel} shows two-dimensional projections of the concept space on MMLU, constructed using the PZS scorer. We illustrate two representative configurations: orthogonal concepts that should exhibit negligible joint activation (abstract algebra vs.\ anatomy, Figure~\ref{fig:mmlu_orthogonal_hexbin}), and related concepts that should co-activate (college physics vs.\ high school physics, Figure~\ref{fig:mmlu_related_hexbin}). Together they show that the induced concept space is interpretable
directly from natural-language requirements.

\subsection{Untagged Corpora: Exploratory Structure}
\label{sec:untagged}

Many real-world text collections are unlabeled; here the goal is
interpretable exploration in concept space rather than recovery of
predefined categories. We illustrate this on a hotel-review corpus
(Sections~\ref{sec:binary-axes}--\ref{sec:graded}) and on UN General
Debate speeches (Section~\ref{sec:un-speeches}).

For the hotel demonstration, we begin from more than 500{,}000
European hotel reviews~\citep{liu2017515k}. To keep the analysis
tractable, we embed reviews with a pretrained sentence
encoder~\citep{reimers2019sentencebertsentenceembeddingsusing}, reduce dimensionality via UMAP~\citep{mcinnes2018umap},
%after a truncated-SVD pre-reduction
and
cluster with HDBSCAN~\citep{mcinnes2017hdbscan}; scoring is performed
on the per-cluster prototype (the review closest to each centroid).
The clustering step is external to ZSV and could be replaced by any
sampling scheme, or skipped entirely when compute permits.

\subsubsection{Binary axes}
\label{sec:binary-axes}
\begin{figure}[!t]
  \centering
  \includegraphics[width=\linewidth]{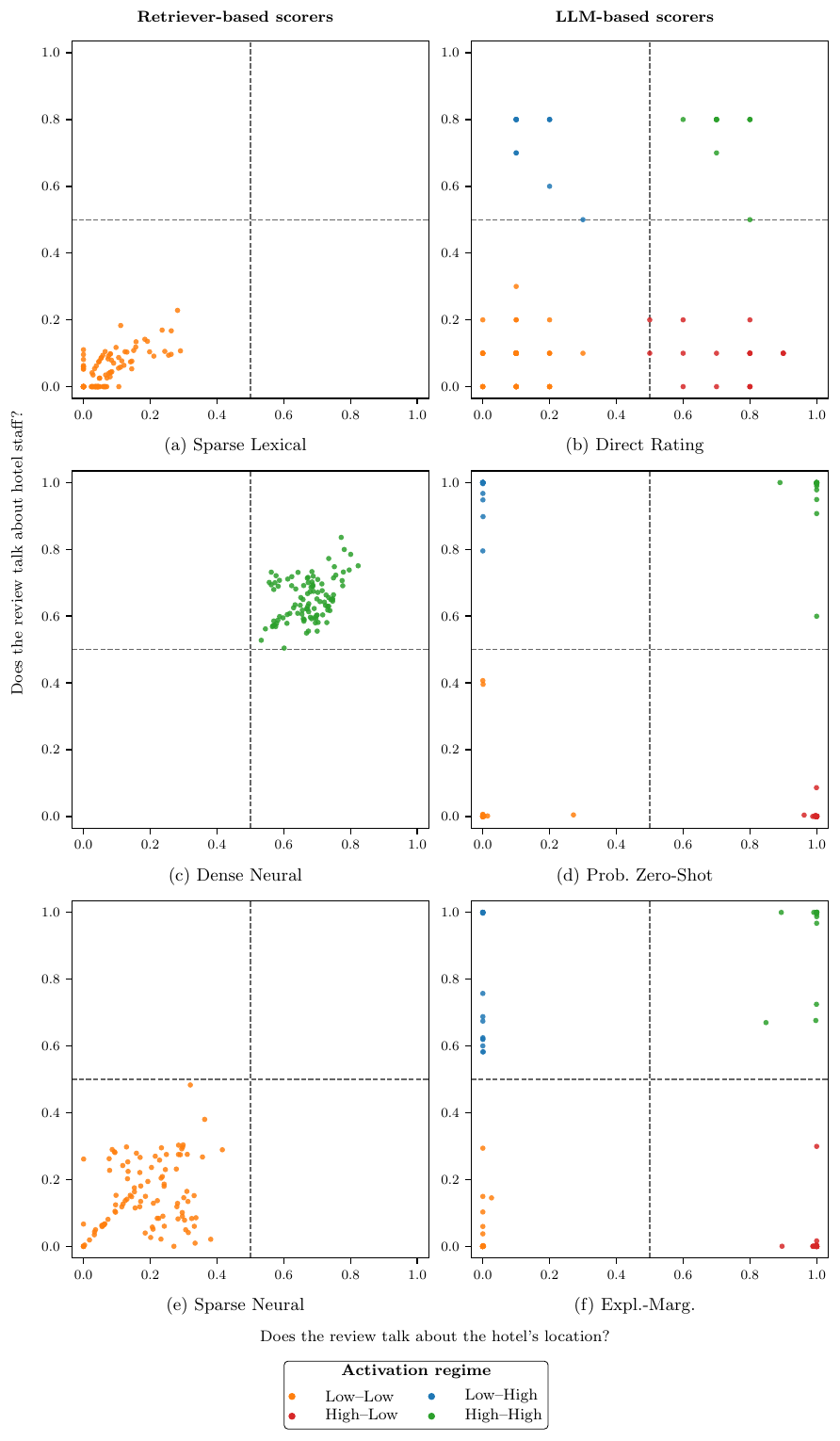}
  \caption{Hotel reviews projected onto the location and staff axes for six scoring methods. Dashed lines mark the $0.5$ thresholds defining the four activation regimes.}
  \label{fig:hotel_multiplot}
\end{figure}

Each prototype $x$ is mapped to a concept activation vector $\mathbf{s}(x)$
over user-defined semantic concepts capturing key aspects of customer
feedback (e.g., staff, location, cleanliness).
Figure~\ref{fig:hotel_multiplot} projects these prototypes onto the staff and location axes for six scoring methods, with a
common coordinate system and a $0.5$ threshold on each axis that partitions
the space into four activation regimes (low--low, high--low,
low--high, high--high).
The two scorer families produce visibly different geometries. Sparse
lexical matching concentrates almost all prototypes in the low--low regime,
while dense neural retrieval collapses them toward the high--high corner;
sparse neural retrieval is slightly better spread but still poorly
differentiated. LLM-based scorers produce markedly more structured layouts.
Direct rating yields a grid-like pattern consistent with discretized
judgments, and the two probabilistic LLM scorers (zero-shot and
explanation-marginalized) deliver the clearest quadrant separation,
consistent with their best ground-truth alignment in Table~\ref{tab:tagged_results} (Brier).
In the PZS projection, the closest example to each corner grounds the
induced geometry in text (e.g.\ ``Excellent staff. Comfortable
rooms. Great location.'' at high--high, ``Breakfast was quite
nice. The rooms were spacious and clean.'' at low--low). The full per-regime summary is in Appendix~\ref{app:hotel_quadrants}.

\subsubsection{Graded axes}
\label{sec:graded}
Binary axes partition a corpus into two groups but discard intensity. We
extend ZSV with \textbf{graded axes}: the binary True/False question is
replaced by an ordinal multiple-choice question whose answer is converted
into a continuous score in $[0,1]$ using the same probability extraction
as PZS (Appendix~\ref{app:graded-prompts}). For instance, the binary
question ``Does this review mention the room?'' becomes
``How satisfied is the reviewer with the room?'' scored on a
five-point Likert scale from ``Very dissatisfied'' to ``Very
satisfied''. This enables fine-grained scatter-plot visualizations of intensity that complement the coarser binary view.

\paragraph{The binary--graded ambiguity.}
Graded axes alone introduce an interpretation problem. A graded score on
``how satisfied is the reviewer with the room?'' is meaningful
only when the review actually discusses the room. When the topic is
absent the scorer has no textual evidence to anchor on and is expected
to fall back toward the middle of the scale, which is indistinguishable
from a review that genuinely expresses a neutral opinion. The two cases
land in the same region of the axis but mean very different things.

\begin{figure}[!t]
  \includegraphics[width=\linewidth]{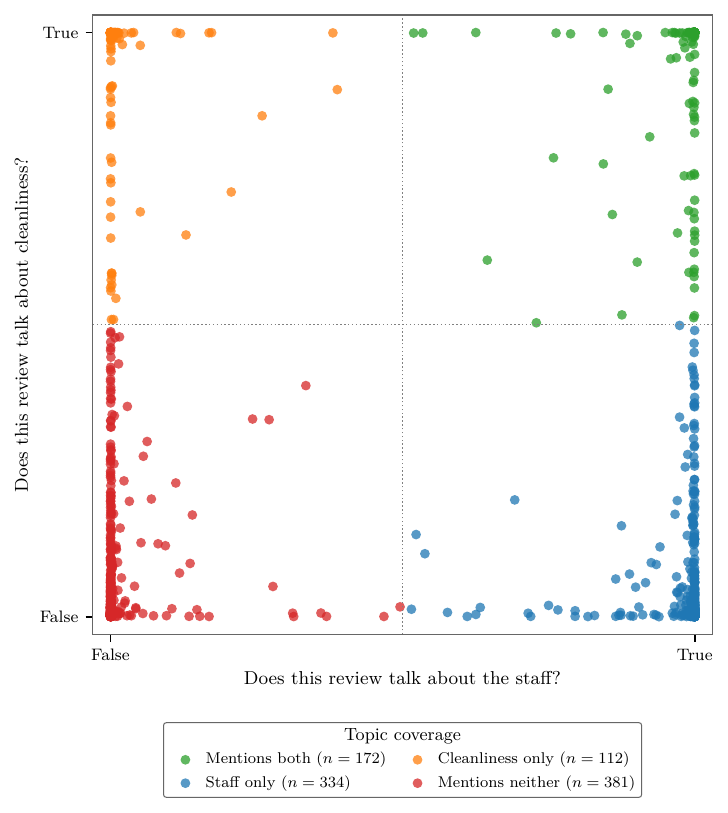}
  \caption{Binary relevance: does each review talk about the staff
  ($x$-axis) and cleanliness ($y$-axis)? The four quadrants partition
  reviews by which topics they actually mention.}
  \label{fig:hotel_relevance_scatter}
\end{figure}

\begin{figure}[!t]
  \centering
  \includegraphics[width=\linewidth]{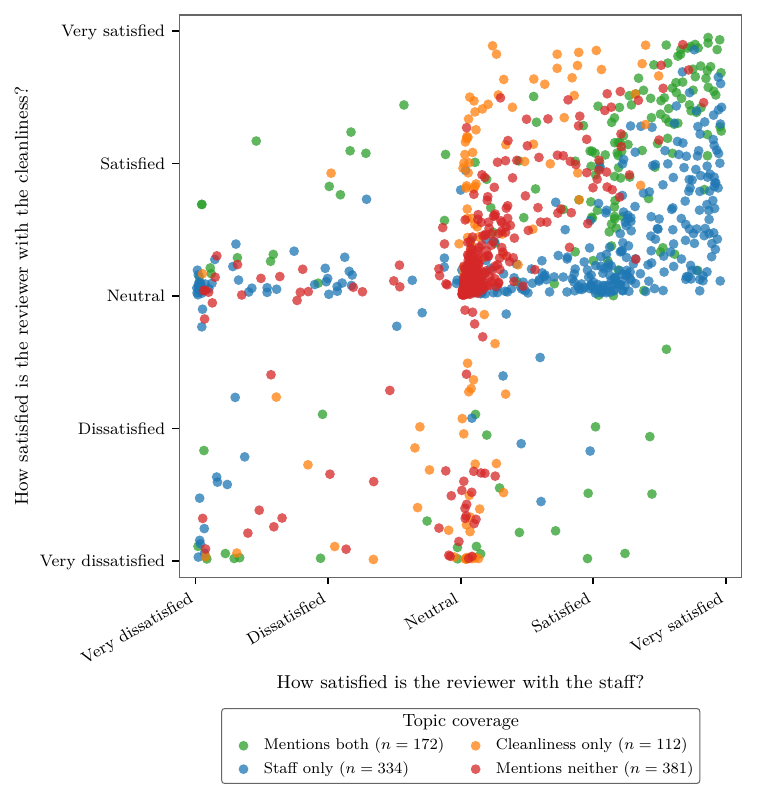}
  \caption{Graded satisfaction with staff ($x$-axis) and cleanliness
  ($y$-axis), coloured by the mention regime from
  Figure~\ref{fig:hotel_relevance_scatter}. The on-topic subset
  (green, both topics mentioned) is shown alone with example reviews
  in the teaser (Figure~\ref{fig:zsv_teaser}).}
  \label{fig:hotel_graded_scatter}
\end{figure}

We resolve the ambiguity by pairing the two scoring modes. A binary
relevance query (``Does this review talk about the staff?'')
decides whether the topic is actually present in the text, and the graded
query measures intensity only where the relevance signal is positive.
Concretely, for two graded axes $(g_x, g_y)$ paired with relevance scores
$(r_x, r_y)$, every review falls into one of four mention regimes:
both topics mentioned, $x$-only, $y$-only, or neither.
Figure~\ref{fig:hotel_relevance_scatter} shows this binary partition for
two example topics; Figure~\ref{fig:hotel_graded_scatter} plots the
corresponding graded scores, coloured by mention regime, potentially showing the analyst
could identify and filter out off-topic documents (indeed, keeping only the ``mentions both'' class yields Figure \ref{fig:zsv_teaser}).

\begin{figure}[!t]
  \centering
  \includegraphics[width=\linewidth]{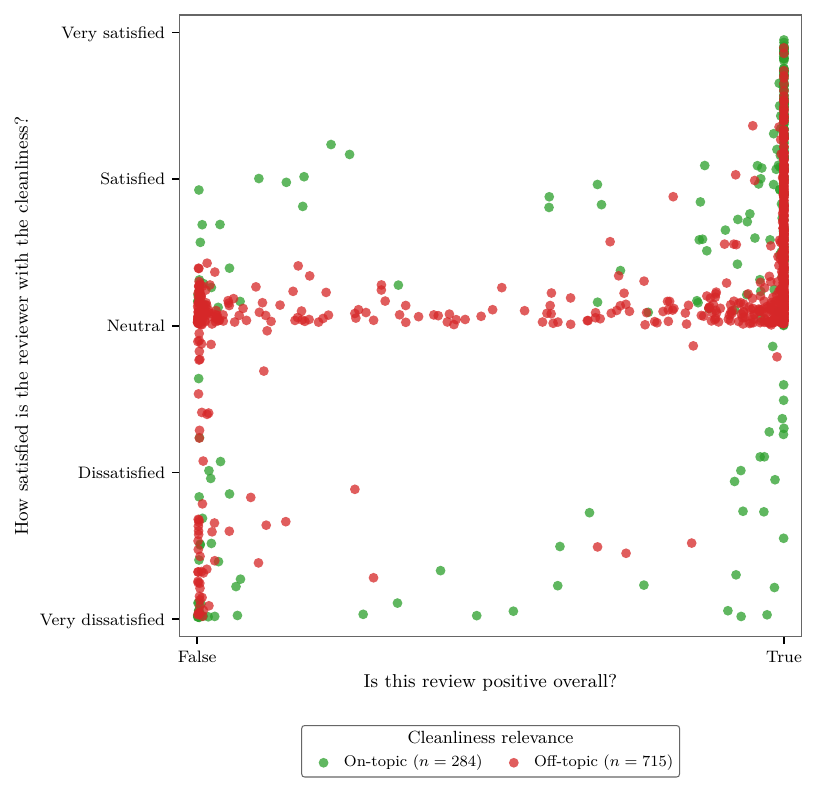}
  \caption{Halo effect on the cleanliness axis. Positive-tone score ($x$) vs cleanliness graded score ($y$). On-topic (green): genuine textual signal. Off-topic (red): should cluster at neutral but instead track sentiment.}
  \label{fig:hotel_halo_panel}
\end{figure}

\paragraph{Compositional sentiment bias.}
Figure~\ref{fig:hotel_halo_panel} plots overall sentiment ($x$) against the graded cleanliness score ($y$). Off-topic reviews (red), which do not actually mention cleanliness, should cluster  at the neutral midpoint of $y$, since the scorer has no textual evidence to anchor on. While the horizontal line is observed, an anomaly appears at the extremes, with inflated cleanliness scores for extremely positive reviews (and, equivalently, deflated scores on the negative side).
Manual inspection confirms that these reviews are indeed off-topic, suggesting that, in the absence of explicit evidence about cleanliness, but under an extreme sentiment, the model seems to ``propagate'' that sentiment to the target concept, similarly to the \textit{halo effect} in the cognitive bias literature \citep{thorndike1920constant,nisbett1977halo}.
We call this {\it compositional sentiment bias}. 

This empirically confirms the need for binary-relevance filtering above to mitigate this bias. In a self-referential sense, the phenomenon was itself uncovered using ZSV, demonstrating the framework’s capacity to reveal unexpected structure in unlabeled data; a full study of this phenomenon is left to future work.

\subsubsection{UN speeches: multi-axis exploration}
\label{sec:un-speeches}

While the hotel demonstration above showcases ZSV on consumer text
with two paired axes, the framework also accommodates non-consumer
corpora and larger axis counts. We illustrate one more example on the
UN General Debate Corpus~\citep{baturo2017understanding}, which
collects the annual address of every member state to the United
Nations General Assembly.

\begin{figure}[!t]
  \centering
  \includegraphics[width=\linewidth]{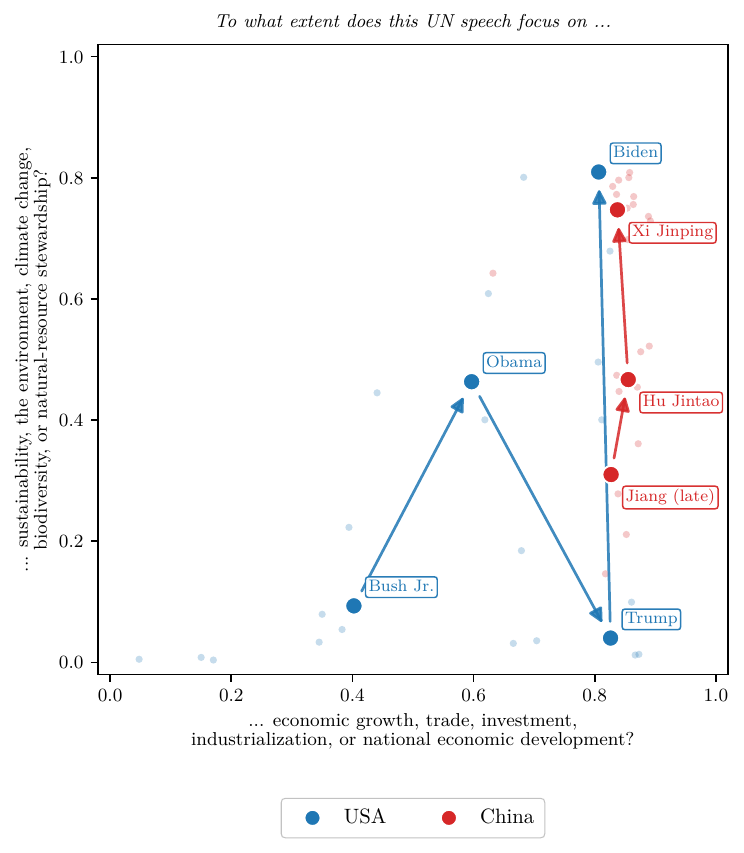}
  \caption{USA and China at the UN General Debate, 2001--2022. Large dots: era centroids by leader, with chronological arrows; small dots: individual speeches.}
  \label{fig:un_trajectory}
\end{figure}

Figure~\ref{fig:un_trajectory} projects each country onto two axes,
aggregated by presidency or leadership era. The induced geometry reveals contrasting, decadal-scale patterns of economic and environmental emphasis between the two countries, again without any fine-tuning or
task-specific training, hinting at ZSV's potential across application domains.

\section{Practitioner Guidance}
\label{sec:practitioner}
\textbf{(1) Default to PZS.} PZS produces the best metrics with reasonable processing times (Table~\ref{tab:tagged_results}).
\textbf{(2) Use DNR when throughput matters more than faithfulness.} DNR offers competitive AUROC while requiring roughly two orders of magnitude less computation  (Table~\ref{tab:tagged_results}). This comes at the price of poorer scaling and ground-truth alignment (Brier \& Figure \ref{fig:hotel_multiplot}).
\textbf{(3) Pick LLMs by Brier, not AUROC.} Across 3B--9B open-weight LLMs (Table~\ref{tab:robustness}), AUROC barely moves while Brier varies sharply; 
\textbf{(4) Pair graded axes with binary-relevance filtering.} Without it, low scores conflate ``not mentioned'' with ``neutral,'' and we observe a compositional sentiment bias  (Section~\ref{sec:graded}).

\section{Conclusion}
\label{sec:conclusion}
We introduced the ZSV task, implementations  for mapping text corpora onto user-defined natural-language axes, and a visualization-first benchmark to evaluate key design choices arising in practice. 

Across tagged and untagged corpora, we showed that probabilistic
zero-shot scoring offers the strongest practical compromise between
semantic faithfulness and computational cost, and that the geometry
induced by different scorer families differs sharply in visualization quality.
Extending ZSV from binary to graded axes makes intensity visible but
introduces a low-score ambiguity (genuinely low versus not mentioned);
pairing each graded axis with a binary relevance query resolves the
ambiguity and yields an interpretable multi-axis visualization. As a
byproduct, the off-regime points expose a sentiment-driven ``halo effect'' in
off-topic scores that we leave for future characterization.

Beyond visualization, the same concept activations can serve as
interpretable features for downstream tasks:
Appendix~\ref{app:surrogates} approximates a high-capacity classifier on
ScienceQA~\citep{lu2022learn} by fitting a shallow decision tree with a few concepts scored with PZS. We provide this experiment to showcase possible future extensions.

\section{Limitations}
\label{sec:limitations}

\paragraph{Scope of evaluation.}
Our quantitative benchmark covers two tagged corpora (MMLU
and Banking77) and one untagged corpus of hotel reviews,
all in English and drawn from general-knowledge,
customer-intent, and consumer-review domains. We do not
evaluate ZSV on multilingual data, specialized domains
such as scientific or legal text, or long-form documents.
The surrogate study in Appendix~\ref{app:surrogates} is
similarly restricted to a single dataset (ScienceQA), one
teacher architecture, and one tree depth.

\paragraph{EM-Hybrid underperformance.}
The EM-Hybrid configuration in
Table~\ref{tab:tagged_results} does not improve over its
components on the evaluated benchmarks. This reflects a
known limit of two-stage retrieval pipelines: when the
first-stage scorer is itself weak, the reranker cannot
recover misranked candidates. We report the result for
completeness rather than as a recommended configuration,
and we do not explore alternative selection policies or
first-stage scorers that might mitigate this behaviour.

\paragraph{Graded axes and halo effect.}
Graded axes are demonstrated only on the untagged hotel
corpus; we do not benchmark them quantitatively against
binary scoring on tagged data, so their semantic
faithfulness relative to PZS is not measured here. The
halo effect we observe in off-topic graded scores is
presented as a qualitative finding surfaced by the
binary--graded visualization workflow, rather than a
systematically characterised phenomenon. Quantifying its
magnitude, and its dependence on scorer and prompt, is left to future work.

\paragraph{Concept design.}
ZSV takes natural-language concepts as exogenous user input: the
interpretability and informativeness of the resulting visualization
depend on how the analyst phrases each concept, and we do not study
concept paraphrase robustness or prompt sensitivity.

\paragraph{Use of AI assistants.}
The authors used a general-purpose large language model
assistant during the preparation of this manuscript for
language polishing, suggesting alternative phrasings,
discussing paper structure and framing, and assisting
with portions of the experimental code. No AI assistant
was used to generate research ideas, design experiments,
run analyses, or produce results. The authors reviewed
and verified all AI-assisted output and take full
responsibility for the content of this paper.

\bibliography{custom}

\appendix

\section{Implementation Details and Prompt Templates}
\label{app:implementation-prompts}
This appendix documents the concrete implementation used to produce all
concept alignment scores reported in the paper, and the exact prompt
templates used by the LLM-based scorers. All scorers output a normalized
alignment score $s(x,c) \in [0,1]$ and differ only in how the alignment
between a text $x$ and a concept $c$ is estimated. Templates are reproduced
verbatim from the implementation; \texttt{\{text\}} and
\texttt{\{requirement\}} are placeholders substituted at scoring time.

\subsection{Decoding and Inference}
\label{app:decoding}

All LLM-based scorers (DR, PZS, and EM) use a single instruction-tuned
open-weight language model, loaded in half precision. Prompts are passed
as raw strings, except in the cross-model comparison
(Table~\ref{tab:robustness}), where each model's own chat template is
applied.

DR generates a single token greedily and parses it as a numeric rating.
PZS, and the entailment and requirement steps of EM, perform a single
forward pass and read the softmax over the full vocabulary at the last
position; no tokens are generated. EM additionally samples short reasoning
traces with $T=0.9$, top-$k=40$, top-$p=0.85$, \texttt{max\_new\_tokens}=30,
producing $N=3$ explanations per (text, requirement) pair.

\subsection{Retriever-Based Scoring}

These scorers use no prompts; alignment is computed in vector space.

\paragraph{Sparse Lexical Retriever (SLR).} Texts and requirements are
encoded as TF--IDF vectors over a shared vocabulary
(\texttt{TfidfVectorizer}, default parameters). Alignment is the cosine
similarity between the resulting sparse vectors.

\paragraph{Dense Neural Retriever (DNR).} Texts and requirements are
encoded with
\texttt{sentence-transformers/all-MiniLM-L6-v2}~\cite{wang2020minilmdeepselfattentiondistillation,reimers2019sentencebertsentenceembeddingsusing}.
Alignment is the cosine similarity between L2-normalized embeddings,
rescaled to $[0,1]$ via $s = (1 + \cos)/2$.

\paragraph{Sparse Neural Retriever (SNR).} Texts and requirements are
encoded with
\texttt{naver/splade-v3}~\cite{lassance2024spladev3newbaselinessplade},
which produces sparse vocabulary-sized vectors with learned term weights.
Alignment is the sparse dot product between text and requirement
representations.

\subsection{LLM-Based Scoring (Binary Mode)}
\label{app:binary-prompts}

\paragraph{Direct Rating (DR).} The model returns a numeric score in
$[0,10]$, linearly mapped to $[0,1]$. If parsing the generated token as a
number fails, the score defaults to $0$.

\begin{quote}\small\ttfamily
You are evaluating how well the following question matches the given topic.\\
\\
Provide a single numeric score from 0 to 10, where:\\
- 0 means the question is completely unrelated to the topic.\\
- 10 means the question is entirely focused on the topic.\\
- Decimal values are allowed if the match is partial.\\
\\
Only return the numeric score. Do not explain your answer.\\
\\
Topic: \{requirement\}\\
Question: \{text\}\\
\\
Score (0--10):
\end{quote}

\paragraph{Probabilistic Zero-Shot (PZS).} A single forward pass yields the
next-token distribution; probability mass for True and False is collected
over the surface variants listed below.

\begin{quote}\small\ttfamily
Text:\\
\{text\}\\
\\
Requirement:\\
\{requirement\}\\
\\
Does the text fulfill the requirement? Respond with either `True' or `False'.\\
Answer:
\end{quote}

The True and False buckets aggregate the following surface forms:

\begin{quote}\small\ttfamily
True:  ``\textvisiblespace True'', ``TRUE'', ``true'', ``True.'', ``true.''\\
False: ``\textvisiblespace False'', ``FALSE'', ``false'', ``False.'', ``false.''
\end{quote}

Each variant is tokenized without special tokens; we keep the first token
id, deduplicate across variants, and sum the probability mass over the
resulting set. The score is
$s(x,c) = p_T / (p_T + p_F + \varepsilon)$ with $\varepsilon = 10^{-6}$.

\paragraph{Explanation-Marginalized (EM).} EM uses three prompts: an
explanation generator, an entailment check, and a requirement check.
Entailment and requirement prompts are scored with the same True/False
mass extraction as PZS, batched over the $N$ explanations.

\textit{Explanation sampling} (sampled $N=3$ times per pair):
\begin{quote}\small\ttfamily
Text: \{text\}\\
Give a short explanation that describes what this text is about and what kind of knowledge it relates to.
\end{quote}

\textit{Entailment check}:
\begin{quote}\small\ttfamily
Text: \{text\}\\
Explanation: \{e\}\\
Is the explanation entailed by the text? Answer with True or False:\\
Answer:
\end{quote}

\textit{Requirement check}:
\begin{quote}\small\ttfamily
Text: \{text\}\\
Explanation: \{e\}\\
Requirement: \{requirement\}\\
Given the text and the explanation, is it reasonable to conclude that the requirement is fulfilled?\\
Answer with True or False:\\
Answer:
\end{quote}

For each explanation $e_i$, let $q_i$ be the True-probability of the
entailment prompt and $a_i$ the True-probability of the requirement
prompt, both computed from the same True/False mass extraction as PZS.
The final score is the entailment-weighted average
$s(x,c) = \sum_i w_i\, a_i$ with weights $w_i = q_i / \sum_j q_j$.

\subsection{Graded Scoring}
\label{app:graded-prompts}

For graded axes, the True/False decision is replaced by a multiple-choice
question over ordinal anchors. The anchor set is part of the axis
definition rather than fixed by the method: the analyst supplies an
ordered list of labels together with their numeric values, so the scale
can be phrased in terms of the concept being measured. All experiments
in this paper use five anchors with values evenly spaced over $[0,1]$
($0$, $0.25$, $0.5$, $0.75$, $1$). Table~\ref{tab:anchors} reports the
label sets used.

\begin{table}[h]
\centering
\small
\begin{tabular}{@{}lp{4.3cm}@{}}
\toprule
Used in & Anchor labels (lowest to highest) \\
\midrule
Fig.~\ref{fig:zsv_teaser}, Fig.~\ref{fig:hotel_graded_scatter} &
Very dissatisfied, Dissatisfied, Neutral, Satisfied, Very satisfied \\
\addlinespace
Fig.~\ref{fig:un_trajectory} &
Not at all, Briefly mentioned, A moderate theme, A major theme,
The dominant theme \\
\addlinespace
Generic default &
Not at all, Slightly, Moderately, Very much, Completely \\
\bottomrule
\end{tabular}
\caption{Ordinal anchor sets used for graded axes. Values are
$0$, $0.25$, $0.5$, $0.75$, $1$ in all cases; only the labels change.}
\label{tab:anchors}
\end{table}

For each option letter $k$, probability mass is summed over the surface
variants \texttt{"A"}, \texttt{"\_A"}, \texttt{"a"}, \texttt{"\_a"},
\texttt{"A)"}, \texttt{"\_A)"}, \texttt{"(A)"}, \texttt{"\_(A)"} (and
analogously for the other letters), then normalized across options. The
graded score is the expected value $\sum_k p_k v_k$ under the resulting
categorical distribution. The prompts below are shown with the generic
default anchors; the anchor labels are substituted at scoring time.

\textit{Graded requirement prompt} (PZS counterpart):
\begin{quote}\small\ttfamily
Text:\\
\{text\}\\
\\
Requirement:\\
\{requirement\}\\
\\
To what extent does the text fulfill the requirement?\\
Options: A) Not at all  B) Slightly  C) Moderately  D) Very much  E) Completely\\
Answer:
\end{quote}

\textit{Graded entailment prompt} (EM counterpart):
\begin{quote}\small\ttfamily
Text: \{text\}\\
Explanation: \{e\}\\
To what extent is the explanation supported by the text?\\
Options: A) Not at all  B) Slightly  C) Moderately  D) Very much  E) Completely\\
Answer:
\end{quote}

\textit{Graded requirement-given-explanation prompt} (EM counterpart):
\begin{quote}\small\ttfamily
Text: \{text\}\\
Explanation: \{e\}\\
Requirement: \{requirement\}\\
Given the text and the explanation, to what extent is the requirement fulfilled?\\
Options: A) Not at all  B) Slightly  C) Moderately  D) Very much  E) Completely\\
Answer:
\end{quote}

\subsection{Hybrid Configuration}

The hybrid scorer pairs SLR as the first-stage scorer with EM ($N=3$)
as the reranker, using top-percent selection per requirement: for each
requirement, the top 10\% of texts under the SLR score are reranked by EM,
while all other (text, requirement) pairs retain their SLR score. The
implementation supports three orthogonal configuration choices that can
be combined freely: the first-stage scorer, the reranker, and the
selection policy (top-$k$, top-percent, or score threshold; applied
per requirement, per text, or globally).

\section{Concept-to-Decision Surrogates}
\label{app:surrogates}
As a side contribution, we apply ZSV beyond visualization to a related interpretability task: testing whether the concept activations can serve as features for downstream prediction.
The concept space induced by ZSV supports interpretability and exploratory
analysis, while also providing a usable feature representation for
downstream decision-making. This appendix probes whether zero-shot concept
activations support prediction without task-specific representation
learning, using \emph{concept-to-decision} surrogates as controlled probes.

\paragraph{Setup (ScienceQA).}
We instantiate this protocol on ScienceQA, where each question is annotated
with a coarse subject label (one of \emph{language}, \emph{natural}, or
\emph{social science}) and a fine-grained topical
category~\citep{lu2022learn}. Concept axes are defined by the fine-grained
categories, mapping each question $x$ to an alignment vector
$\mathbf{s}(x) \in [0,1]^K$ scored with PZS.

\paragraph{Teacher and surrogate distillation.}
To isolate representational quality from classifier capacity, we adopt a
two-stage teacher-surrogate distillation
setup~\citep{NIPS1995_45f31d16,hinton2015distillingknowledgeneuralnetwork,frosst2017distillingneuralnetworksoft}.
First, a high-capacity \textit{teacher} model is trained on the \texttt{train}
split to predict subject probabilities from an embedding representation of
the text. Second, an interpretable \textit{surrogate} decision tree is trained
on the \texttt{validation} split using only the concept vectors
$\mathbf{s}(x)$ as input features and the teacher's predicted probability
vectors as training targets. The surrogate is constrained to be shallow
(\texttt{max\_depth}=3), ensuring compact decision paths composed entirely
of thresholded, human-readable concept tests
(Figure~\ref{fig:surrogate-decision-tree}).

\paragraph{Evaluation.}
Both teacher and surrogate are evaluated on the held-out \texttt{test}
split. We report (i) \textit{surrogate fidelity}, measured by agreement with
the teacher's probabilistic outputs, and (ii) \textit{task performance}
against ground-truth subject labels. Results are summarized in
Table~\ref{tab:surrogate-fidelity}. To make the induced decision logic concrete, Table~\ref{tab:surrogate-examples} presents one correct prediction and one misclassification for which the surrogate faithfully reproduces the teacher's wrong answer, illustrating how decision paths support both interpretation of correct decisions and post-hoc diagnosis of model errors.

\begin{table}[!t]
\centering
\footnotesize
\setlength{\tabcolsep}{5pt}
\renewcommand{\arraystretch}{1.1}
\begin{tabular}{lc}
\toprule
\textbf{Metric (test)} & \textbf{Value} \\
\midrule
\multicolumn{2}{l}{\textbf{Surrogate fidelity (teacher $\rightarrow$ surrogate)}} \\
MSE & 0.062 \\
$R^2$ & 0.662 \\
Agreement & 0.832 \\
\midrule
\multicolumn{2}{l}{\textbf{Task accuracy (vs.\ ground truth)}} \\
Teacher accuracy & 0.988 \\
Surrogate accuracy & 0.826 \\
\bottomrule
\end{tabular}
\caption{Surrogate fidelity and task accuracy on ScienceQA.}
\label{tab:surrogate-fidelity}
\end{table}

\begin{figure}[!t]
\centering
 
\begin{tikzpicture}[
    font=\footnotesize,
>=Latex,
    level distance=14mm,
    level 1/.style={sibling distance=42mm},
    level 2/.style={sibling distance=20mm},
    level 3/.style={sibling distance=9mm},
    every node/.style={draw, rounded corners=3pt, align=center, inner sep=3pt, font=\scriptsize},
    split/.style={fill=gray!10, text width=1.7cm},
    leaf/.style={fill=gray!5, text width=0.6cm, font=\scriptsize\bfseries},
    edge from parent/.style={draw, -Latex}
]
 
% -------- Header --------
\node[draw=none, fill=none, font=\footnotesize] (hdr)
{All internal nodes test: \emph{``Is this text about \texttt{<concept>}?''}};
 
% ---------------- Root ----------------
\node[split, below=4mm of hdr] (root) {
  \texttt{geography}\\
  $\le 0.994$
}
% -------- Level 1 --------
child { node[split] {
\texttt{literacy-in-}\\
\texttt{science}\\
$\le 0.044$
}
    % -------- Level 2 (left subtree) --------
    child { node[split] {
    \texttt{figurative-}\\
    \texttt{language}\\
    $\le 0.010$
    }
        % -------- Leaves --------
        child { node[leaf] {\textbf{Nat.}} }
        child { node[leaf] {\textbf{Lang.}} }
    }
    child { node[split] {
    \texttt{grammar}\\
    $\le 0.521$
    }
        child { node[leaf] {\textbf{Nat.}} }
        child { node[leaf] {\textbf{Lang.}} }
    }
}
% -------- Level 1 (right subtree) --------
child { node[split] {
\texttt{us-history}\\
$\le 0.004$
}
    child { node[leaf] {\textbf{Nat.}} }
    child { node[split] {
    \texttt{civics}\\
    $\le 0.010$
    }
        child { node[leaf] {\textbf{Soc.}} }
        child { node[leaf] {\textbf{Soc.}} }
    }
};

% ---- Manual Yes/No label near top-left edge ----
\node[
  draw,
  rounded corners=2pt,
  fill=white,
  inner sep=1.2pt,
  font=\scriptsize
] at ([xshift=-13.5mm,yshift=-5.5mm]root) {True};

\node[
  draw,
  rounded corners=2pt,
  fill=white,
  inner sep=1.2pt,
  font=\scriptsize
] at ([xshift=13.5mm,yshift=-5.5mm]root) {False};
 
\end{tikzpicture}

\caption{
Decision-tree surrogate operating on zero-shot concept activations. Internal
nodes threshold human-defined requirements, and leaves output the predicted
ScienceQA subject (\textbf{Nat.}, \textbf{Soc.}, \textbf{Lang.}).
}
 
\label{fig:surrogate-decision-tree}
\end{figure}
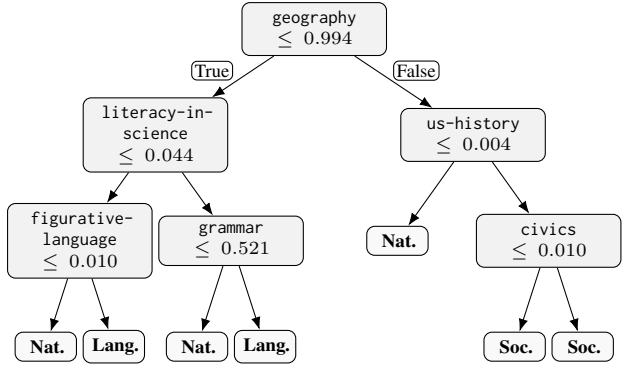
\begin{table}[!t]
\centering
\small
\setlength{\tabcolsep}{3pt}
\renewcommand{\arraystretch}{1.15}
\begin{tabular}{p{0.95\linewidth}}
\toprule
\textbf{Illustrative surrogate explanations} \\
\midrule

\midrule
\textbf{\uppercase{Correct classification}} \\
\midrule
\textbf{Text:} Can \emph{Fromia monilis} cells make their own food? \\
\textbf{True subject:} Natural science \\
\textbf{Decision path:} \\
``Is this text about \ldots'' \\
\texttt{geography}? $s = 0.0004 \le \tau = 0.994$ \\
\quad$\rightarrow$ \texttt{literacy-in-science}? $s = 0.993 > \tau = 0.044$ \\
\qquad$\rightarrow$ \texttt{grammar}? $s = 0.002 \le \tau = 0.521$ \\
\qquad\qquad$\Rightarrow$ \textbf{Predicted: Natural science} \\

\midrule
\textbf{\uppercase{Misclassification}} \\
\midrule
\textbf{Text:} In which place are cows, horses, and other livestock more common? \\
\textbf{True subject:} Social science \\
\textbf{Teacher prediction:} Natural science ($p = 0.78$) \\
\textbf{Decision path:} \\
``Is this text about \ldots'' \\
\texttt{geography}? $s = 0.981 \le \tau = 0.994$ \\
\quad$\rightarrow$ \texttt{literacy-in-science}? $s = 0.482 > \tau = 0.044$ \\
\qquad$\rightarrow$ \texttt{grammar}? $s = 0.003 \le \tau = 0.521$ \\
\qquad\qquad$\Rightarrow$ \textbf{Predicted: Natural science} \\
\textbf{Traceable cause:} the surrogate reproduces the teacher's error rather
than masking it, which is the desired behaviour for a faithful explanation.
The decision path makes the error legible: the root \texttt{geography} concept
is highly activated but falls just below the threshold
($s = 0.981 < \tau = 0.994$), routing the example to a subtree without
\emph{Social science} leaves. References to animals and livestock then make the
\emph{Natural science} prediction a plausible semantic confusion. The surrogate
thus exposes a concrete failure mechanism of the underlying model: a single
brittle threshold separating geography from non-geography concepts.
\\

\bottomrule
\end{tabular}
\caption{
Surrogate decision paths for a correct and an incorrect classification, showing
alignment scores $s(x,c)$, thresholds $\tau$, and error sources.
}
\label{tab:surrogate-examples}
\end{table}

\section{Hotel Quadrant Examples}
\label{app:hotel_quadrants}
Table~\ref{tab:hotel_quadrants_examples_app} reports cluster counts
per activation regime and the prototype closest to each corner, for
the PZS projection in Figure~\ref{fig:hotel_multiplot}.

\begin{table}[h]
\centering
\footnotesize
\setlength{\tabcolsep}{3pt}
\renewcommand{\arraystretch}{1.15}
\begin{tabularx}{\linewidth}{lcX}
\toprule
\textbf{Quadrant} & \textbf{\# Clusters} & \textbf{Representative prototype} \\
\midrule
Low--Low   & 30 & Breakfast was quite nice. The rooms were spacious and clean. \\
Low--High  & 18 & The front desk staff was nice and helpful. \\
High--Low  & 36 & The location was fantastic. The bed was comfortable. \\
High--High & 19 & Excellent staff. Comfortable rooms. Great location. \\
\bottomrule
\end{tabularx}
\caption{Cluster-level summary for the PZS projection in
Figure~\ref{fig:hotel_multiplot}.}
\label{tab:hotel_quadrants_examples_app}
\end{table}

\end{document}